\documentclass[11pt]{article}

\usepackage[preprint]{acl}

\usepackage{times}
\usepackage{latexsym}

\usepackage[T1]{fontenc}
\usepackage[utf8]{inputenc}
\usepackage{microtype}
\usepackage{inconsolata}

\usepackage{graphicx}
\usepackage{multirow}
\usepackage{amsmath}
\usepackage{algorithm}
\usepackage{algpseudocode}
\usepackage{soul}
\usepackage{xcolor}
\usepackage{eso-pic}
\usepackage{booktabs}
\usepackage{amsfonts}
\usepackage{nicefrac}
\usepackage{colortbl}
\usepackage{xspace}
\usepackage{bm}
\usepackage{enumitem}
\usepackage{subcaption}
\usepackage{pifont}

\definecolor{darkgreen}{RGB}{0,128,0}

\newcommand{\methodname}{\textsc{DEAR}\xspace}

\newcommand{\piS}{\pi_\theta}
\newcommand{\piT}{\pi_T}
\newcommand{\entropy}{\mathcal{H}}
\newcommand{\divg}{\delta}
\newcommand{\attn}{\hat{a}}
\newcommand{\divhat}{\hat{\delta}}
\newcommand{\cossim}{\mathrm{cos}}
\newcommand{\respond}{\mathbf{y}}
\newcommand{\prompt}{\mathbf{x}}
\newcommand{\hiddenstate}{\mathbf{h}}

\usepackage[most,skins,theorems]{tcolorbox}

\usepackage{amsthm}
\usepackage{amssymb}
\theoremstyle{plain}

\theoremstyle{remark}

\AddToShipoutPictureBG{%
  \ifnum\value{page}=1
    \AtPageUpperLeft{%
      \put(\LenToUnit{2.5cm},\LenToUnit{-2.5cm}){%
        \includegraphics[height=0.8cm]{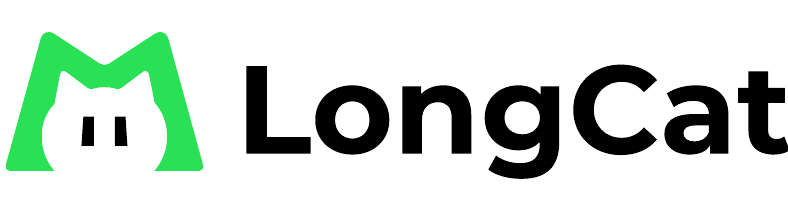}%
      }%
    }%
  \fi
}

\title{Finding the Evidence: Discovering Decision-Supporting Tokens\\for On-Policy Reasoning Distillation}

\author{
 \textbf{Jinwei Xiao\textsuperscript{1}},
 \textbf{Zhuowen Han\textsuperscript{3}},
 \textbf{Yueqing Sun\textsuperscript{1}},
 \textbf{Zhengxi Lu\textsuperscript{1}},
 \textbf{Yuxin Liu\textsuperscript{1}},
\\
 \textbf{Zhiyuan Yao\textsuperscript{1}},
 \textbf{Wentao Chen\textsuperscript{2}}, 
 \textbf{Qi Gu$^{\dagger}$\textsuperscript{1}},
 \textbf{Xunliang Cai\textsuperscript{1}},
\\
\\
 \textsuperscript{1}Meituan Longcat Team,
 \textsuperscript{2}Nanjing University,
\\
 \textsuperscript{3}TJUNLP Lab, College of Intelligence and Computing, Tianjin University
\\
 \texttt{\small xiaojinwei0917@gmail.com\qquad guqi03@meituan.com} 
}

\begin{document}
\maketitle

\begin{abstract}
On-policy distillation transfers reasoning ability through dense token-level supervision, yet the nature of the transferable signal remains unclear.
We discover that reasoning chains contain two types of knowledge that require different discovery mechanisms: \emph{decisions} (where to branch), which surface through student uncertainty, and \emph{evidence} (intermediate steps that justify decisions), which hides in positions where the student is confident yet wrong.
Current methods capture only decisions; the substantive knowledge in evidence tokens remains untransferred.
We propose \methodname (\textbf{D}ecision-\textbf{E}vidence \textbf{A}ware \textbf{R}easoning Distillation), which first identifies decisions via student entropy, then discovers their supporting evidence through hidden-state cosine similarity to decision anchors, boosted by teacher--student divergence to prioritize the largest knowledge gaps.
Across three student--teacher configurations on math and code benchmarks, \methodname consistently outperforms standard OPD, with up to +2.5pp on competition math and +5.7pp on code generation.
\end{abstract}

\section{Introduction}
\label{sec:intro}

\begin{figure*}[t!]
\centering
\includegraphics[width=\textwidth]{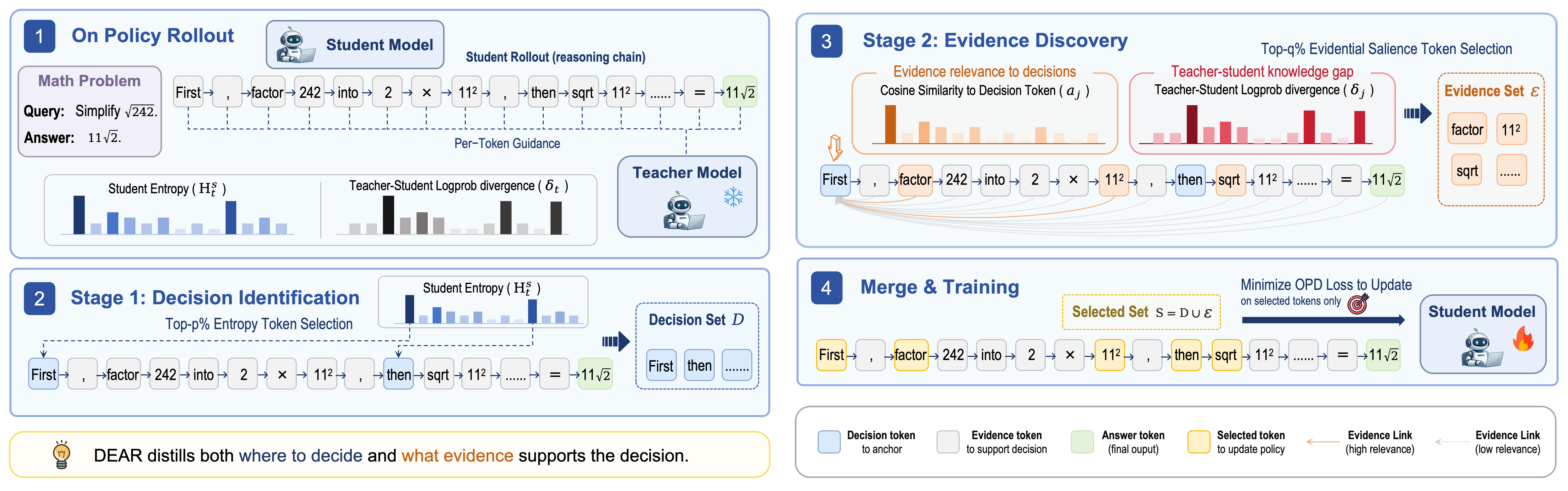}
\caption{\textbf{Overview of \methodname.} \ding{182}~The student generates a rollout; both student entropy $\entropy_t^S$ and teacher--student logprob divergence $\divg_t$ are computed per token. \ding{183}~Decision Identification: the top-$p$\% entropy tokens form the decision set $\mathcal{D}$. \ding{184}~Evidence Discovery: non-decision tokens are scored by cosine similarity to decisions, boosted by divergence; the top-$q$\% form the evidence set $\mathcal{E}$. \ding{185}~The OPD loss is computed only on $\mathcal{S} = \mathcal{D} \cup \mathcal{E}$ to update the policy of student model.}
\label{fig:method}
\end{figure*}

On-policy distillation (OPD) has become the dominant paradigm for teaching small models to reason: the student generates trajectories under its own policy, and a teacher provides dense token-level supervision~\citep{agarwal2024policy,gu2024minillm}.
Despite its effectiveness, OPD leaves a critical question open: what should the student learn from each token?
Training on all tokens dilutes the signal; recent work selects high-entropy tokens to focus learning~\citep{wang2026beyond}, yet the distilled student still often fails on substance.
It writes ``Let us try factoring'' at the right moment, but the intermediate step is wrong, the variable substitution is incorrect, the derivation does not follow.

A closer look reveals that tokens carrying reasoning ability fall into two distinct categories (\S\ref{sec:analysis_motivation}):
\begin{itemize}[nosep,leftmargin=*]
    \item \textbf{Decision tokens}: positions where the student faces genuine uncertainty among reasoning paths, such as logical connectives, strategy choices, and branching markers. These surface naturally through student entropy.
    \item \textbf{Evidence tokens}: positions where the student confidently produces the \emph{wrong} intermediate result, such as incorrect derivations, flawed variable substitutions, or erroneous API calls. These are invisible to entropy because the student does not know what it does not know.
\end{itemize}
These two types cannot be discovered by the same mechanism.
Entropy exposes the former but is blind to the latter.
Current training methods, whether full-token or entropy-selective~\citep{wang2026beyond,ko2026scaling,jin2026entropy}, either dilute the signal across all tokens or capture only decisions, leaving the substantive reasoning knowledge untransferred (\S\ref{sec:evidence_tokens}).

How can evidence be discovered?
The key insight is that genuine evidence tokens share reasoning context with nearby decisions: an incorrect intermediate result participates in the same derivation as the decision point that initiated it.
In contrast, noise sources (error accumulation, stylistic divergence) lack this connection (\S\ref{sec:evidence_context}).

We propose \methodname (\textbf{D}ecision-\textbf{E}vidence \textbf{A}ware \textbf{R}easoning Distillation), a two-stage token selection method for OPD.
Stage~1 identifies decisions via student entropy.
Stage~2 discovers evidence through hidden-state cosine similarity to decision anchors, boosted by teacher--student divergence to prioritize the largest knowledge gaps.
Our contributions are:
\begin{itemize}[nosep,leftmargin=*]
    \item We identify a decomposition of reasoning chains into decisions and evidence, two types of knowledge that require fundamentally different discovery mechanisms (\S\ref{sec:analysis_motivation}).
    \item We propose \methodname, which discovers evidence via its relationship to decisions: cosine similarity ensures reasoning-context relatedness, and divergence boost prioritizes knowledge gaps.
    \item On six math benchmarks and three code benchmarks across three student--teacher configurations, \methodname outperforms standard OPD by up to +2.5pp on competition math and +5.7pp on code generation (\S\ref{sec:experiments}).
\end{itemize}

\section{Dissecting the Distillation Signal}
\label{sec:analysis_motivation}

We analyze the token-level distillation signal to uncover the relationship between reasoning decisions and their supporting evidence.
All analyses use rollout data from the first training epoch of Qwen2.5-1.5B-Instruct~\citep{DBLP:journals/corr/abs-2412-15115} (student) and Qwen3-4B-Instruct~\citep{DBLP:journals/corr/abs-2505-09388} (teacher) on DeepMath-103K~\citep{DBLP:journals/corr/abs-2504-11456}, with $n{=}200$ samples.
This cross-family pair (Setting~B in \S\ref{sec:setup}) maximizes the decision--evidence gap, making the phenomenon most visible.

\begin{tcolorbox}[colback=green!5, colframe=green!5, boxrule=0pt, arc=1.5pt, left=4pt, right=4pt, top=2pt, bottom=2pt]
\textbf{Finding 1.} The distillation signal is extremely sparse. A small minority of tokens carry the vast majority of gradient mass, motivating token selection.
\end{tcolorbox}
\label{sec:sparsity}

\begin{figure}[htb]
\centering
\begin{subfigure}[t]{0.48\columnwidth}
    \centering
    \includegraphics[width=\linewidth]{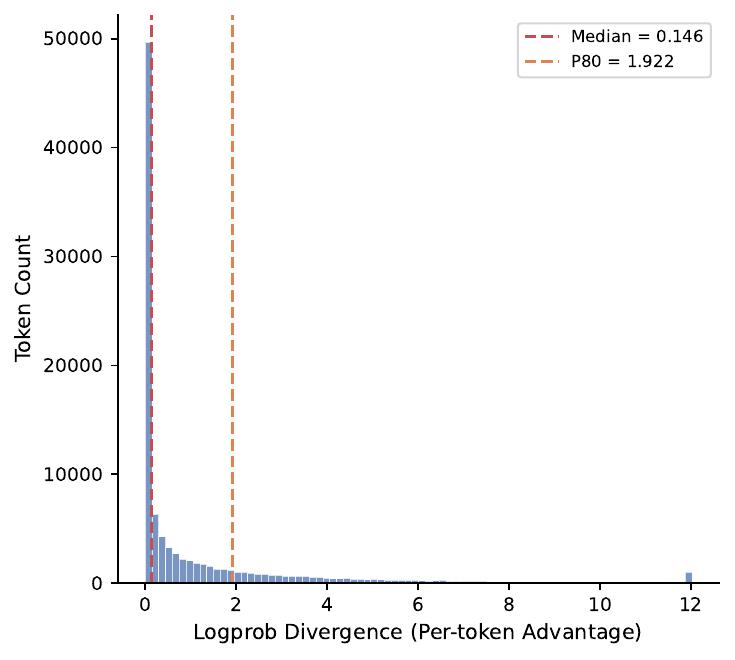}
    \caption{Logprob divergence distribution}
\end{subfigure}
\hfill
\begin{subfigure}[t]{0.48\columnwidth}
    \centering
    \includegraphics[width=\linewidth]{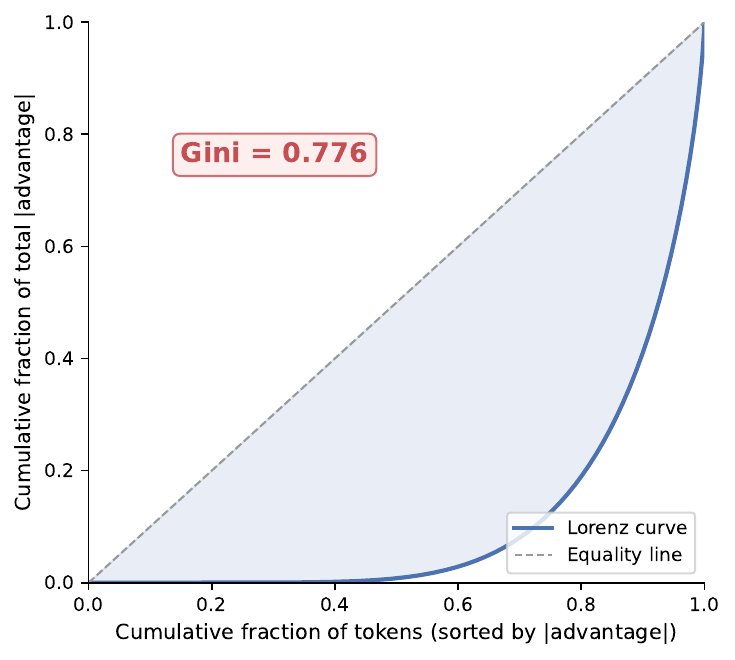}
    \caption{Lorenz curve (Gini = 0.776)}
\end{subfigure}
\caption{\textbf{The distillation signal is extremely sparse.} The top-20\% of tokens carry ${\sim}$80\% of total gradient mass (Gini = 0.776).}
\label{fig:lorenz}
\end{figure}

We measure signal concentration using the Lorenz curve, which plots cumulative gradient mass against the cumulative fraction of tokens sorted by per-token advantage magnitude.
A perfectly uniform signal would follow the diagonal; deviation indicates inequality, quantified by the Gini coefficient (0 = uniform, 1 = maximally concentrated).
Figure~\ref{fig:lorenz} shows a Gini of 0.776: the top-20\% of tokens account for approximately 80\% of total gradient mass.
This extreme sparsity motivates token selection, but the critical question is not how many tokens to select, but what kind.

\begin{tcolorbox}[colback=green!5, colframe=green!5, boxrule=0pt, arc=1.5pt, left=4pt, right=4pt, top=2pt, bottom=2pt]
\textbf{Finding 2.} Entropy selects the reasoning \emph{skeleton}, not the reasoning \emph{knowledge}. High-entropy positions are connectives and branching markers; substantive intermediate steps are uniformly low-entropy.
\end{tcolorbox}
\label{sec:anatomy_decisions}

\begin{figure}[t!]
\centering
\includegraphics[width=0.95\columnwidth]{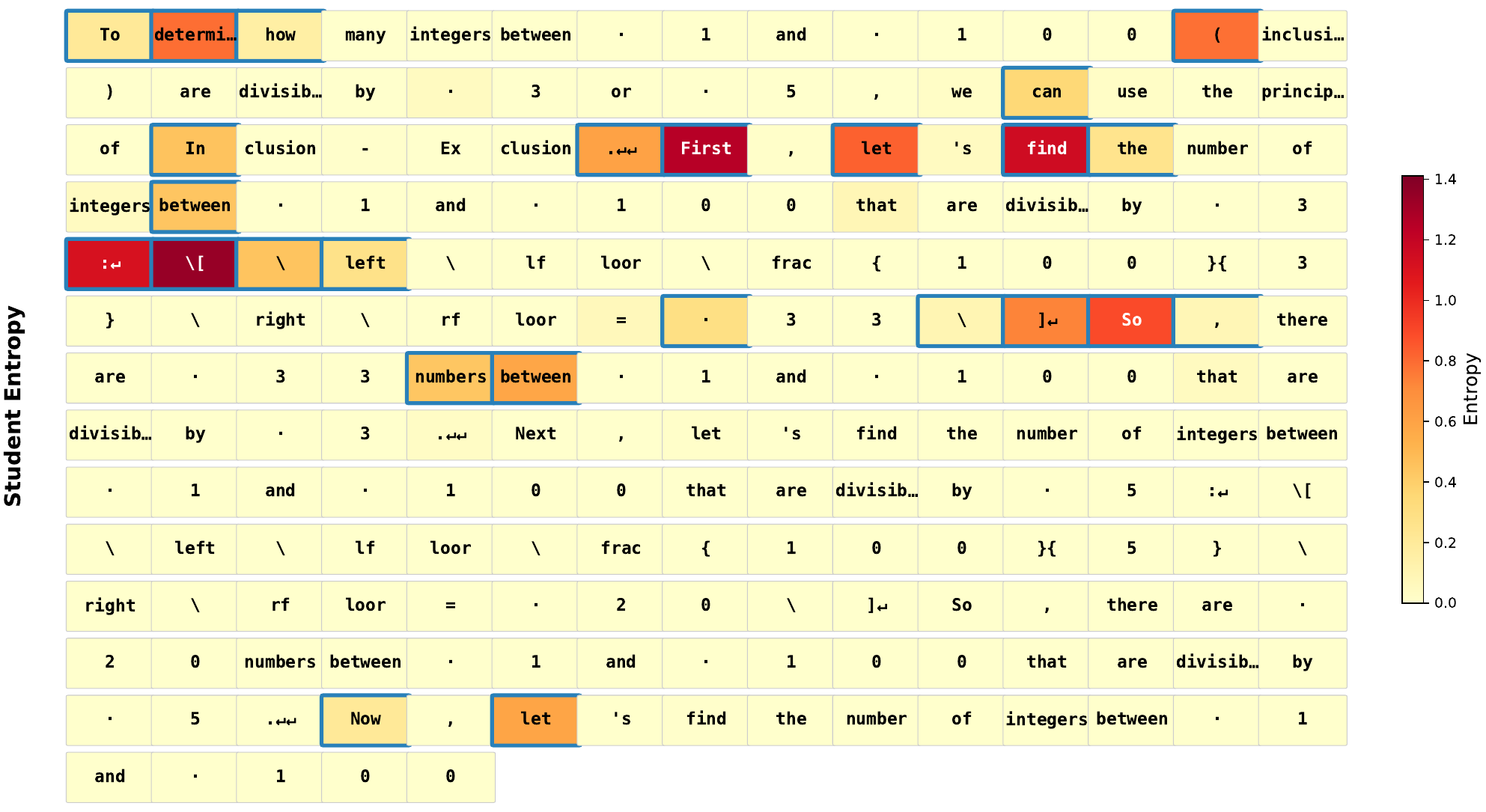}
\caption{\textbf{Entropy concentrates on the reasoning skeleton.} Bright positions align with logical connectives and branching markers; intermediate-step tokens are uniformly dark regardless of correctness.}
\label{fig:entropy_heatmap}
\end{figure}

Figure~\ref{fig:entropy_heatmap} visualizes per-token entropy on a representative math problem.
Intermediate-step tokens, including those where the student writes an \emph{incorrect} value, remain uniformly low-entropy.

Entropy selection therefore captures where the student decides but not what it produces in between.
This explains its effectiveness~\citep{wang2026beyond}: decision tokens are genuinely critical for training.
Yet it also exposes a gap: the reasoning knowledge that supports those decisions lives in low-entropy positions that no entropy-based selector can reach.

\begin{tcolorbox}[colback=green!5, colframe=green!5, boxrule=0pt, arc=1.5pt, left=4pt, right=4pt, top=2pt, bottom=2pt]
\textbf{Finding 3.} Missed reasoning knowledge concentrates in ``evidence tokens'' where the student is confident yet wrong. These positions are structurally invisible to entropy selection.
\end{tcolorbox}
\label{sec:evidence_tokens}

\begin{figure}[t!]
\centering
\includegraphics[width=0.95\columnwidth]{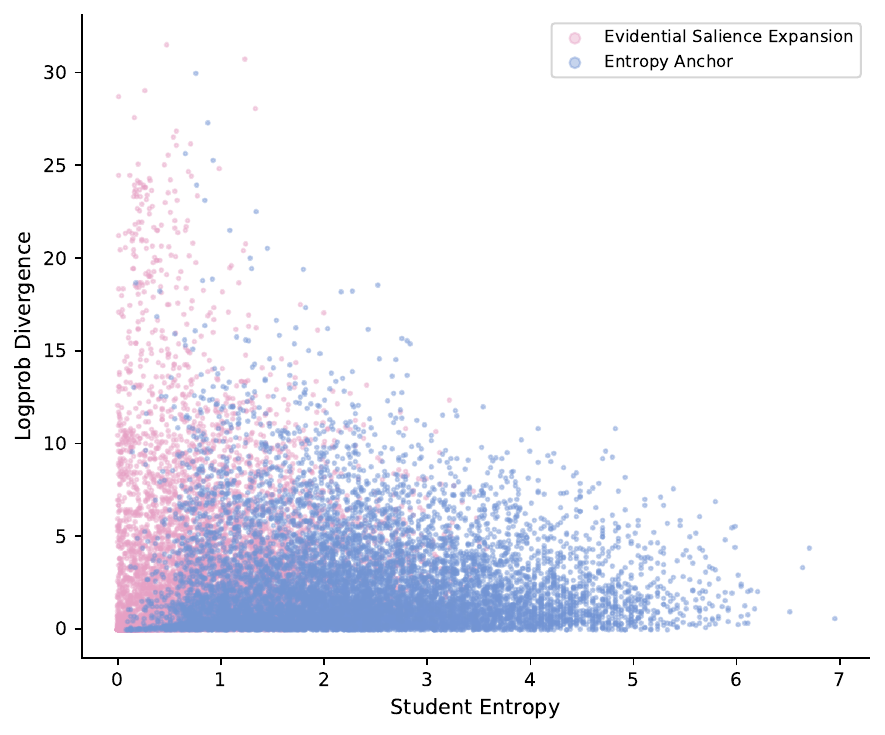}
\caption{\textbf{The token landscape.} ``Entropy Anchor'' tokens (blue) are captured by entropy selection. ``Evidential Salience Expansion'' tokens (pink) concentrate in the low-entropy, high-divergence region.}
\label{fig:scatter}
\end{figure}

Where does the missed knowledge reside?
We plot all response tokens in the $(\entropy_t^S, \divg_t)$ plane (Figure~\ref{fig:scatter}), where $\divg_t = |\log \piS(y_t) - \log \piT(y_t)|$ measures teacher--student disagreement.
A distinct cluster appears in the upper-left: low student entropy paired with high divergence.
These are positions where the student confidently writes a wrong value that the teacher would correct.

We term these \textbf{evidence tokens}.
As we show in \S\ref{sec:exp_analysis}, they carry a substantial fraction of total gradient mass yet achieve near-zero recall under entropy selection.
This is not a tuning failure but a structural impossibility: any monotone function of student entropy necessarily assigns near-zero importance to low-entropy positions, regardless of their divergence.

\begin{tcolorbox}[colback=green!5, colframe=green!5, boxrule=0pt, arc=1.5pt, left=4pt, right=4pt, top=2pt, bottom=2pt]
\textbf{Finding 4.} Genuine evidence tokens share reasoning context with nearby decisions; noise tokens with similar divergence do not.
\end{tcolorbox}
\label{sec:evidence_context}

\begin{figure}[t!]
\centering
\includegraphics[width=0.95\columnwidth]{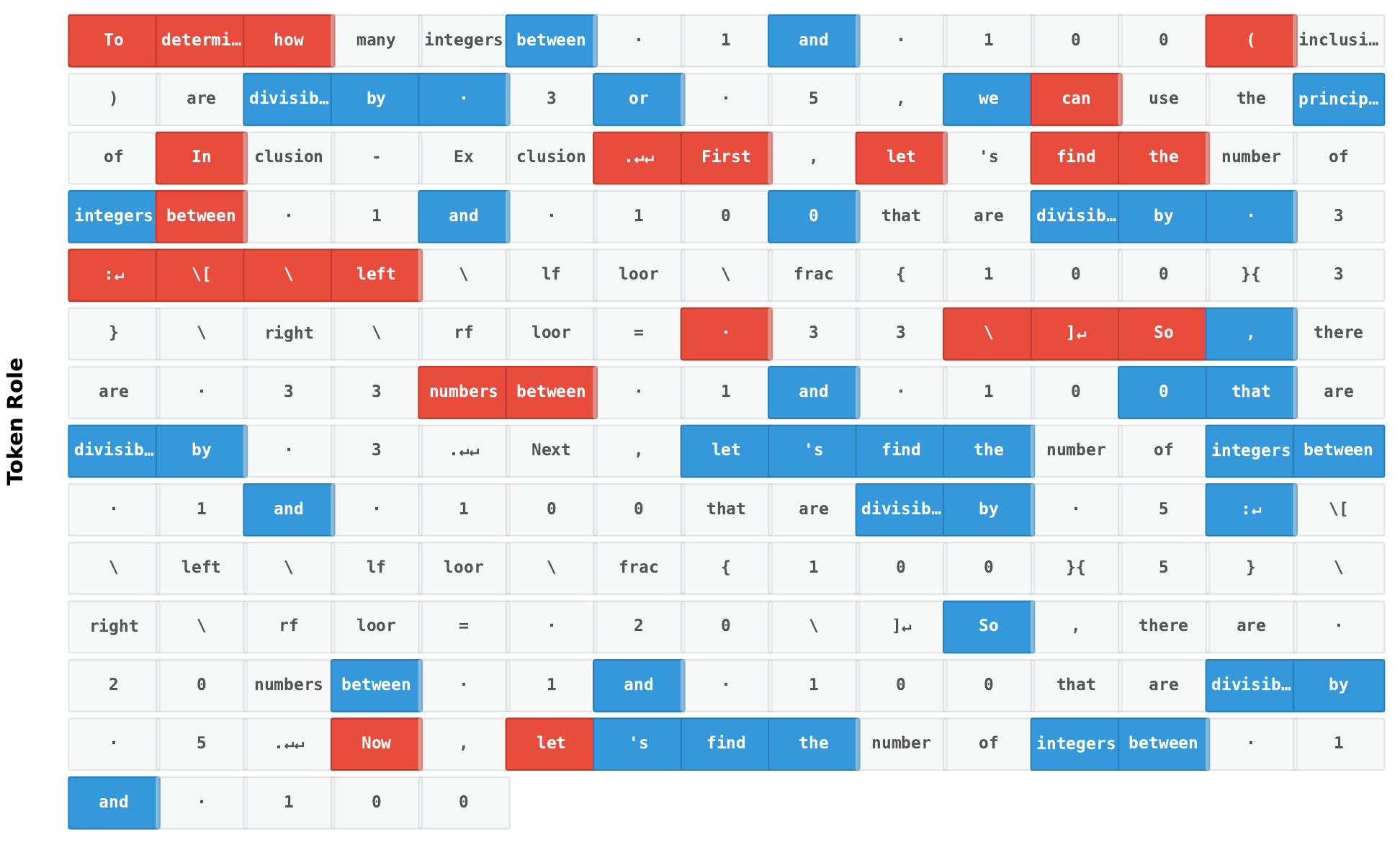}
\caption{\textbf{Evidence clusters around decisions.} \textcolor{red}{\textbf{Decision}} tokens (red) mark reasoning branches; \textcolor{blue}{\textbf{Evidence}} tokens (blue) perform intermediate steps within the same reasoning chain. Unselected tokens (gray) are disconnected from any decision.}
\label{fig:qualitative}
\end{figure}

Selecting by divergence alone would recover evidence but also introduce noise: error accumulation~\citep{li2026rethinking} and stylistic differences both produce high divergence without supporting any decision.
Figure~\ref{fig:qualitative} shows the key distinction: genuine evidence clusters around decisions, while noise tokens appear in isolation.

We operationalize this via hidden-state cosine similarity: a token's maximum similarity to any decision anchor measures whether it participates in the same reasoning context, motivating cosine similarity as the evidence-detection gate.

\section{Method}
\label{sec:method}

Our analysis reveals that complete reasoning distillation requires capturing both decision tokens and evidence tokens.
Entropy selection identifies decisions; our task is to discover their supporting evidence.
We now formalize this intuition.

\subsection{Background}
\label{sec:background}

\paragraph{On-policy distillation.}
Let $\piS$ denote the student policy parameterized by $\theta$, and $\piT$ a frozen teacher.
Given a prompt $\prompt$, the student samples a response $\respond = (y_1, \dots, y_T)$.
PG-style OPD~\citep{yang2026learning,ko2026scaling} computes a per-token advantage from the teacher's log-probabilities:
\begin{equation}
    A_t = -\bigl(\log \piS(y_t \mid \prompt, y_{<t}) - \log \piT(y_t \mid \prompt, y_{<t})\bigr),
    \label{eq:advantage}
\end{equation}
which arises from differentiating the sequence-level reverse KL divergence $\mathcal{D}_{\mathrm{KL}}(\piS \| \piT)$ (the full derivation is provided in Appendix~\ref{app:opd_derivation}), and optimizes a clipped policy-gradient loss:
\begin{equation}
\label{eq:opd_loss}
\begin{split}
    \mathcal{L}_{\text{OPD}} = -\frac{1}{T}\sum_{t=1}^{T} &\mathrm{sg}(A_t) \cdot \min\bigl(\rho_t \cdot \mathrm{sg}(A_t),\\
    &\mathrm{clip}(\rho_t, 1{-}\epsilon, 1{+}\epsilon)\cdot \mathrm{sg}(A_t)\bigr),
\end{split}
\end{equation}
where $\rho_t = \piS(y_t) / \piS^{\text{old}}(y_t)$ is the importance ratio and $\mathrm{sg}(\cdot)$ denotes stop-gradient.

\paragraph{Entropy-selective OPD (decision-only distillation).}
The student's per-token entropy is:
\begin{equation}
    \entropy_t^S = -\sum_v \piS(v \mid \prompt, y_{<t}) \log \piS(v \mid \prompt, y_{<t}).
    \label{eq:entropy}
\end{equation}
Entropy selection~\citep{wang2026beyond} restricts the loss to the top-$p$\% highest-entropy positions, which, as established in \S\ref{sec:anatomy_decisions}, correspond to reasoning decision tokens.
Let $\tau_p$ denote the $(1{-}p)$-th quantile of $\{\entropy_t^S\}_{t=1}^{T}$.
The decision set is:
\begin{equation}
    \mathcal{D} = \{t : \entropy_t^S \geq \tau_p\}.
    \label{eq:decision_set}
\end{equation}
This captures the reasoning skeleton but cannot reach evidence tokens (\S\ref{sec:evidence_tokens}).

\subsection{\methodname: Decision-Evidence Aware Reasoning Distillation}
\label{sec:dear}

\methodname extends decision-only distillation by discovering the evidence tokens that support each decision (Figure~\ref{fig:method}).

\paragraph{Stage 1: Decision identification.}
The top-$p$\% tokens by student entropy are identified as decision tokens $\mathcal{D}$ (Eq.~\ref{eq:decision_set}).
These mark where the student faces genuine reasoning choices.
They serve as both training targets and query points for evidence discovery.

\paragraph{Stage 2: Evidence discovery.}
Given a set of decisions, we ask: \emph{which other tokens support those decisions?}
We extract the student's hidden state $\hiddenstate_t^L \in \mathbb{R}^d$ at the deepest transformer layer $L$ from the forward pass already used to compute entropy.
For each non-decision token $y_j$, we compute its maximum cosine similarity to any decision:
\begin{equation}
    a_j = \max_{i \in \mathcal{D}}\; \frac{\hiddenstate_i^L \cdot \hiddenstate_j^L}{\|\hiddenstate_i^L\| \cdot \|\hiddenstate_j^L\|}.
    \label{eq:cosine}
\end{equation}
High $a_j$ means that position $j$ shares reasoning context with a decision, participating in the same reasoning step or producing a related intermediate result.
Since each hidden state $\hiddenstate_t^L$ compresses all causally accessible context into a next-token prediction~\citep{elhage2021mathematical}, cosine similarity in this space captures \emph{reasoning-context sharing} rather than surface structural similarity (see Appendix~\ref{app:cosine_analysis}).
This allows evidence discovery to propagate from decisions to the intermediate steps that support them.

\paragraph{Knowledge-gap scoring.}
Not all tokens related to a decision need correction; some evidence the student already knows.
To select evidence where the teacher has knowledge the student lacks, we incorporate teacher--student divergence as a boost.
Both signals are normalized to $[0, 1]$ via per-sample min-max:
\begin{equation}
    \attn_j = \mathrm{MinMax}(a_j),\quad \divhat_j = \mathrm{MinMax}(\divg_j),
    \label{eq:normalize}
\end{equation}
and combined as Evidence Salience Score:
\begin{equation}
    s_j = \underbrace{\attn_j}_{\text{evidence relevance}} \times \underbrace{(1 + \divhat_j)}_{\text{knowledge gap}}.
    \label{eq:hybrid}
\end{equation}
The cosine similarity $\attn_j$ acts as the evidence gate: tokens semantically distant from all decisions receive near-zero scores regardless of divergence.
As established in \S\ref{sec:evidence_context}, noise sources (error accumulation, style) lack semantic connection to decisions, so the gate naturally excludes them.
The divergence $\divhat_j$ then boosts among evidence candidates, prioritizing those where the teacher has knowledge the student lacks.
We ablate this scoring design in Appendix~\ref{app:scoring_ablation}.

\paragraph{Final selection: decisions $\cup$ evidence.}
We select the top-$q$\% non-decision tokens by hybrid score and take their union with decisions:
\begin{equation}
    \mathcal{S} = \underbrace{\mathcal{D}}_{\text{decisions}} \;\cup\; \underbrace{\mathrm{TopK}_q\!\bigl(\{s_j\}_{j \notin \mathcal{D}}\bigr)}_{\text{evidence}}.
    \label{eq:selection}
\end{equation}
The OPD loss (Eq.~\ref{eq:opd_loss}) is computed only over $\mathcal{S}$, which forms the complete distillable reasoning chain: both where to decide and what justifies those decisions.
With $p{=}0.2$ and $q{=}0.2$, the total selection covers approximately 36\% of response tokens.

\paragraph{Computational overhead.}
The hidden states $\hiddenstate_t^L$ are extracted from the student's existing forward pass (\texttt{output\_hidden\_states=True}), requiring no additional forward pass.
Evidence discovery reduces to a single batched matrix multiplication $\bar{\mathbf{H}}_{\mathcal{D}} \!\cdot\! \bar{\mathbf{H}}_{\text{all}}^\top$ with cost $\mathcal{O}(|\mathcal{D}| \!\times\! T \!\times\! d)$, negligible compared to the transformer forward pass.
The complete procedure is summarized in Algorithm~\ref{alg:dear}.

\begin{algorithm}[t]
\small
\setlength{\abovedisplayskip}{2pt}
\setlength{\belowdisplayskip}{2pt}
\caption{\methodname: Finding Evidence for Reasoning Decisions}
\label{alg:dear}
\algrenewcommand{\algorithmiccomment}[1]{\hfill{\footnotesize\textcolor{gray!80}{$\triangleright$ #1}}}
\begin{algorithmic}[1]
\Require Student $\piS$, teacher $\piT$, prompt $\prompt$, ratios $p, q$
\Ensure Selected token mask $\mathcal{S}$
\State $\respond \sim \piS(\cdot \mid \prompt)$
\Statex \hfill{\footnotesize\textcolor{gray!80}{$\triangleright$ Student rollout}}
\State Compute $\log \piS(y_t)$, $\entropy_t^S$, $\hiddenstate_t^L$ $\forall t$
\Statex \hfill{\footnotesize\textcolor{gray!80}{$\triangleright$ Student forward (with hidden states)}}
\State Compute $\log \piT(y_t)$ $\forall t$
\Statex \hfill{\footnotesize\textcolor{gray!80}{$\triangleright$ Teacher forward}}
\State $I \gets |\log \piS(y_t) - \log \piT(y_t)|$ $\forall t$
\Statex \hfill{\footnotesize\textcolor{gray!80}{$\triangleright$ Per-token knowledge gap}}
\Statex \hfill{\footnotesize\textcolor{gray!80}{$\triangleright$ \textit{Stage 1: Decision Identification}}}
\State $\mathcal{D} \gets \{t : \entropy_t^S \geq \tau_p\}$
\Statex \hfill{\footnotesize\textcolor{gray!80}{$\triangleright$ Top-$p$\% entropy $\to$ decisions}}
\Statex \hfill{\footnotesize\textcolor{gray!80}{$\triangleright$ \textit{Stage 2: Evidence Discovery}}}
\State $\bar{\hiddenstate}_t \gets \hiddenstate_t^L / \|\hiddenstate_t^L\|$ $\forall t$
\Statex \hfill{\footnotesize\textcolor{gray!80}{$\triangleright$ $\ell_2$ normalize}}
\State $a_j \gets \max_{i \in \mathcal{D}} \bar{\hiddenstate}_i \!\cdot\! \bar{\hiddenstate}_j$ $\forall j \notin \mathcal{D}$
\Statex \hfill{\footnotesize\textcolor{gray!80}{$\triangleright$ Evidence relevance to decisions}}
\State $\attn_j, \divhat_j \gets \mathrm{MinMax}(a_j), \mathrm{MinMax}(\divg_j)$
\Statex \hfill{\footnotesize\textcolor{gray!80}{$\triangleright$ Normalize to $[0,1]$}}
\State $s_j \gets \attn_j \times (1 + \divhat_j)$ $\forall j \notin \mathcal{D}$
\Statex \hfill{\footnotesize\textcolor{gray!80}{$\triangleright$ Evidence relevance $\times$ knowledge gap}}
\State $\mathcal{E} \gets \mathrm{TopK}_q(\{s_j\}_{j \notin \mathcal{D}})$
\Statex \hfill{\footnotesize\textcolor{gray!80}{$\triangleright$ Top-$q$\% evidence tokens}}
\State \Return $\mathcal{S} \gets \mathcal{D} \cup \mathcal{E}$
\Statex \hfill{\footnotesize\textcolor{gray!80}{$\triangleright$ Decisions $\cup$ Evidence}}
\end{algorithmic}
\end{algorithm}

\section{Experiments}
\label{sec:experiments}

\subsection{Setup}
\label{sec:setup}

\paragraph{Benchmarks.}
We evaluate on six mathematical reasoning benchmarks: MATH-500~\citep{DBLP:conf/iclr/Gao0YCMDLMCXTWZ25}, Minerva Math~\citep{lewkowycz2022solving}, AMC 2023, Olympiad Bench~\citep{DBLP:conf/acl/HeLBHTSHHHZLQL024}, AIME 2024~\citep{aime24}, and AIME 2025~\citep{aime25}.
We report Avg@8: for each problem, we sample 8 responses at temperature 1.0 and report the average accuracy across samples.
We also evaluate on code generation (\S\ref{sec:code_results}).

\paragraph{Baselines.}
We compare: (1)~\textbf{Offline KD}, SFT on teacher-generated solutions; (2)~\textbf{Standard OPD}, training on all response tokens; and (3)~\textbf{\methodname} with $p{=}0.2$ decision ratio and $q{=}0.2$ evidence ratio.
We additionally compare against \textbf{decision-only distillation} (entropy-selective OPD~\citep{wang2026beyond}) in our ablation study (\S\ref{sec:ablation}), where removing Stage~2 reduces \methodname to this baseline.
All methods train for the same epochs over the same prompt set.

\paragraph{Model configurations.}
We test three student--teacher pairs to vary the decision--evidence gap.
\textbf{Setting~A}: Qwen2.5-1.5B-Instruct~\citep{DBLP:journals/corr/abs-2412-15115} as student distilled from Qwen2.5-14B-Instruct (same family, moderate gap).
\textbf{Setting~B}: Qwen2.5-1.5B-Instruct as student distilled from Qwen3-4B-Instruct~\citep{DBLP:journals/corr/abs-2505-09388} (cross-family, large gap).
\textbf{Setting~C}: Qwen3-1.7B as student distilled from Qwen3-4B-Instruct (same family, stronger student, smaller gap).
Settings A and B share the same student, isolating the effect of decision--evidence alignment.
Training details are in Appendix~\ref{app:impl_details}.

\subsection{Main Results: Mathematical Reasoning}
\label{sec:main_results}

\begin{table*}[t]
\centering
\caption{\textbf{Main results on mathematical reasoning benchmarks.} Avg@8 accuracy (\%). Best per-benchmark in \textbf{bold}, second-best \underline{underlined}.}
\label{tab:main}
\resizebox{\textwidth}{!}{%
\begin{tabular}{ll cccccc}
\toprule
\textbf{Setting} & \textbf{Method} & \textbf{MATH-500} & \textbf{Minerva} & \textbf{AMC23} & \textbf{Olympiad} & \textbf{AIME25} & \textbf{AIME24} \\
\midrule
\multirow{4}{*}{\shortstack[l]{(A) s: Qwen2.5-1.5B \\ t: Qwen2.5-14B}}
& Base model     & 43.33 & 9.10  & 19.06 & 12.44 & 0.00  & 1.25  \\
& Offline KD     & 46.17 & \textbf{14.57} & 23.13 & 16.77 & \underline{0.83} & 1.25  \\
& Standard OPD   & \underline{51.78} & 12.73 & \underline{26.25} & \underline{17.14} & 0.00  & \underline{2.08}  \\
& \methodname    & \textbf{53.15} & \underline{13.83} & \textbf{29.06} & \textbf{18.53} & \textbf{1.25}  & \textbf{4.58}  \\
\midrule
\multirow{4}{*}{\shortstack[l]{(B) s: Qwen2.5-1.5B \\ t: Qwen3-4B}}
& Base model     & 43.68 & 8.69  & 20.63 & 12.57 & 0.42  & 1.67  \\
& Offline KD     & 43.85 & 12.82 & 23.75 & 14.97 & 0.42  & 2.08  \\
& Standard OPD   & \underline{55.40} & \underline{11.63} & \textbf{32.50} & \underline{20.31} & \underline{1.25} & \textbf{5.42}  \\
& \methodname    & \textbf{56.20} & \textbf{13.60} & \underline{30.94} & \textbf{21.05} & \textbf{3.33} & \underline{4.17}  \\
\midrule
\multirow{4}{*}{\shortstack[l]{(C) s: Qwen3-1.7B \\ t: Qwen3-4B}}
& Base model     & 71.18 & 20.31 & 46.88 & 35.00 & 12.50 & 14.58 \\
& Offline KD     & 80.27 & \textbf{33.23} & 61.56 & \underline{49.63} & 20.83 & 22.92 \\
& Standard OPD   & \underline{83.45} & 27.94 & \underline{69.06} & 49.20 & \underline{23.75} & \underline{31.25} \\
& \methodname    & \textbf{84.35} & \underline{28.68} & \textbf{71.25} & \textbf{50.10} & \textbf{25.83} & \textbf{34.18} \\
\bottomrule
\end{tabular}%
}
\end{table*}

Table~\ref{tab:main} reports results across all three configurations.
\methodname achieves the best or second-best result on every benchmark across all settings, with the largest gains concentrating on the hardest problems.

\paragraph{On-policy methods generally outperform offline KD.}
Offline KD improves over the base model but is surpassed by on-policy methods on most benchmarks.
The exception is Minerva Math, where offline KD's advantage may stem from its shorter, more formulaic solutions that suit the benchmark format.
Overall, the gap favoring on-policy methods is largest in Setting~B, where off-policy teacher trajectories poorly match the student's generation style.

\paragraph{Evidence discovery amplifies gains on the hardest problems.}
\methodname's improvement concentrates on competition-level benchmarks where multi-step derivations create long evidence chains.
On AIME~2024, \methodname achieves 4.58\% vs.\ 2.08\% for standard OPD in Setting~A, a 2.2$\times$ improvement.
On AIME~2025 in Setting~B, \methodname reaches 3.33\% vs.\ 1.25\%, a 2.7$\times$ improvement.
Harder problems produce more intermediate steps per decision, meaning more evidence tokens to discover and more knowledge to transfer.

\paragraph{Cross-family distillation: larger knowledge gap, larger gain.}
Setting~B (cross-family) shows the largest absolute gain on AIME benchmarks.
When student and teacher come from different model families, the teacher possesses reasoning knowledge that is maximally different from the student's internalized patterns, making evidence discovery especially valuable.

\subsection{Generalization to Code Generation}
\label{sec:code_results}

To evaluate whether \methodname generalizes beyond mathematical reasoning, we conduct experiments on code generation using Setting~A (Qwen2.5-1.5B-Instruct distilled from Qwen2.5-14B-Instruct).
We use Eurus-RL-Code~\citep{DBLP:journals/corr/abs-2502-01456} (25K samples) as training data and evaluate on three benchmarks: MBPP+~\citep{DBLP:conf/nips/LiuXW023}, HumanEval~\citep{chen2021evaluating}, and APPS~\citep{DBLP:conf/nips/HendrycksBKMAGB21}.

\begin{table}[t]
\centering
\small
\caption{\textbf{Generalization to code generation}. Pass rate (\%). Best in \textbf{bold}, second-best \underline{underlined}.}
\label{tab:code}
\begin{tabular}{l ccc}
\toprule
\textbf{Method} & \textbf{MBPP} & \textbf{HumanEval} & \textbf{APPS} \\
\midrule
Base model      & \underline{53.36} & 47.56 & 10.51 \\
Offline KD      & 43.24 & 45.43 & 5.03 \\
Standard OPD    & 51.31 & \underline{47.71} & \underline{14.34} \\
\methodname     & \textbf{57.05} & \textbf{49.84} & \textbf{16.90} \\
\bottomrule
\end{tabular}
\end{table}

Table~\ref{tab:code} shows that \methodname's gains transfer to code generation.
Offline KD degrades below the base model on all benchmarks, confirming that off-policy trajectories are harmful when execution correctness demands precise token-level alignment.
Standard OPD also degrades MBPP , consistent with the noise amplification problem~\citep{luo2026demystifying}.
\methodname resolves both issues: it outperforms standard OPD by +5.74pp on MBPP, +2.13pp on HumanEval, and +2.56pp on APPS.
In code, decision tokens correspond to control-flow choices while evidence tokens are the API calls and implementation steps that realize those choices.

\subsection{Analysis: Closing the Loop}
\label{sec:exp_analysis}

We analyze \methodname to verify that the discovered tokens are indeed evidence tokens that support decisions, and to understand why evidence-enriched training helps.

\paragraph{Evidence tokens carry disproportionate learning signal.}

\begin{figure}[t!]
\centering
\begin{subfigure}[t]{0.48\columnwidth}
    \centering
    \includegraphics[width=\linewidth]{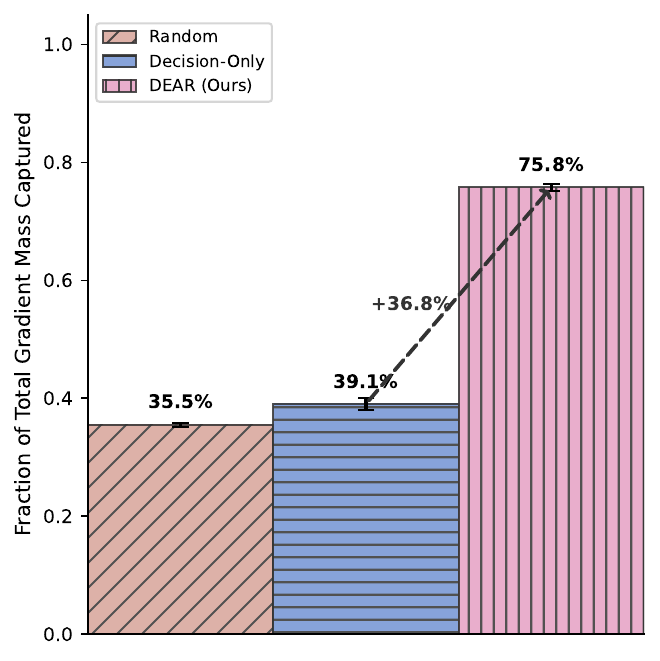}
    \caption{Gradient mass captured}
    \label{fig:recall}
\end{subfigure}
\hfill
\begin{subfigure}[t]{0.48\columnwidth}
    \centering
    \includegraphics[width=\linewidth]{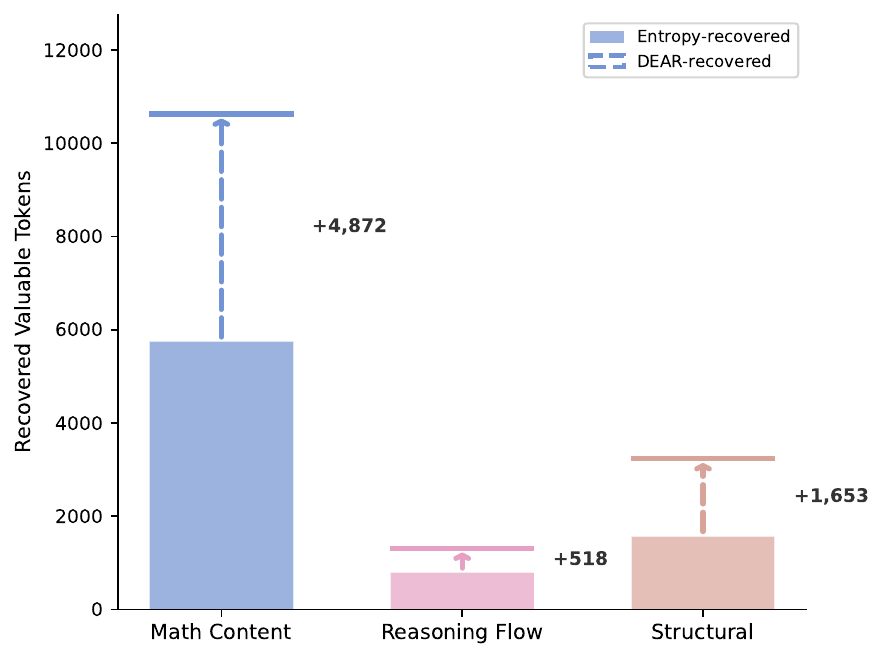}
    \caption{Recovery by category}
    \label{fig:recovery}
\end{subfigure}
\caption{\textbf{Evidence discovery targets high-value tokens.} (a)~Gradient mass captured by each method. (b)~Recovery count stratified by semantic category.}
\label{fig:gradient_recovery}
\end{figure}

Figure~\ref{fig:recall} quantifies the coverage gap: decision-only selection (entropy) captures 39.1\% of total gradient mass, only marginally above a random baseline (35.9\%). Adding evidence discovery via \methodname raises coverage to 75.8\%, nearly doubling the captured signal.

\paragraph{Closing the loop: decisions are skeleton, evidence is substance.}
\begin{figure}[t!]
\centering
\includegraphics[width=0.85\columnwidth]{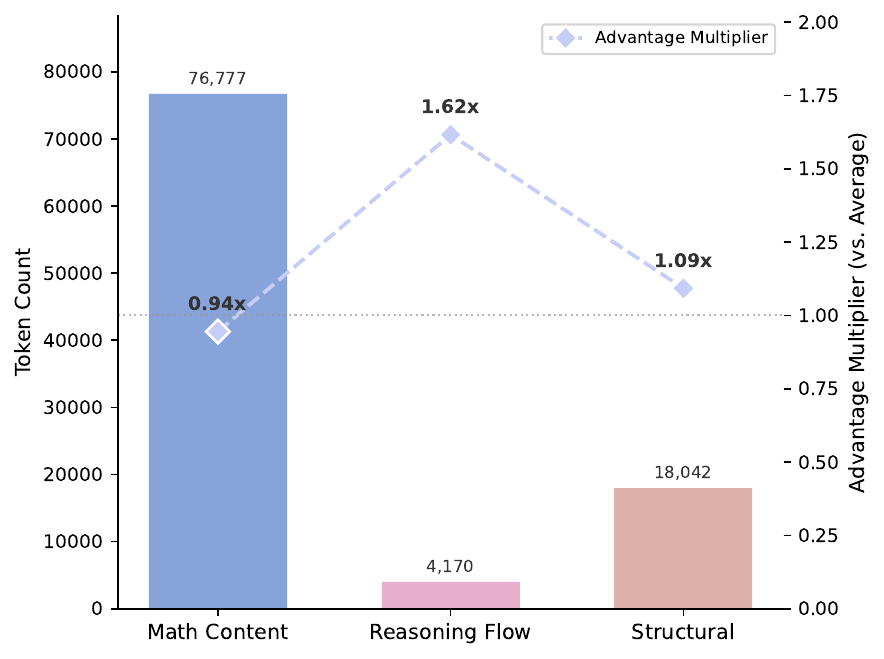}
\caption{\textbf{Token category distribution and advantage multiplier.} Bars show token counts; dashed line shows per-category advantage relative to the per-token average.}
\label{fig:semantic}
\end{figure}

We classify all response tokens by semantic category (Figure~\ref{fig:semantic}).
Reasoning Flow tokens, though rare at only 4,170 occurrences, carry 1.62$\times$ the per-token average advantage, confirming that decisions concentrate disproportionate signal.
Math Content tokens number 76,777 and carry 0.94$\times$ individually, yet dominate total gradient mass by volume.
This confirms the complementary roles: entropy captures decisions, while \methodname recovers the numerous evidence positions where the knowledge gap is large.

\begin{figure}[t!]
\centering
\includegraphics[width=0.95\columnwidth]{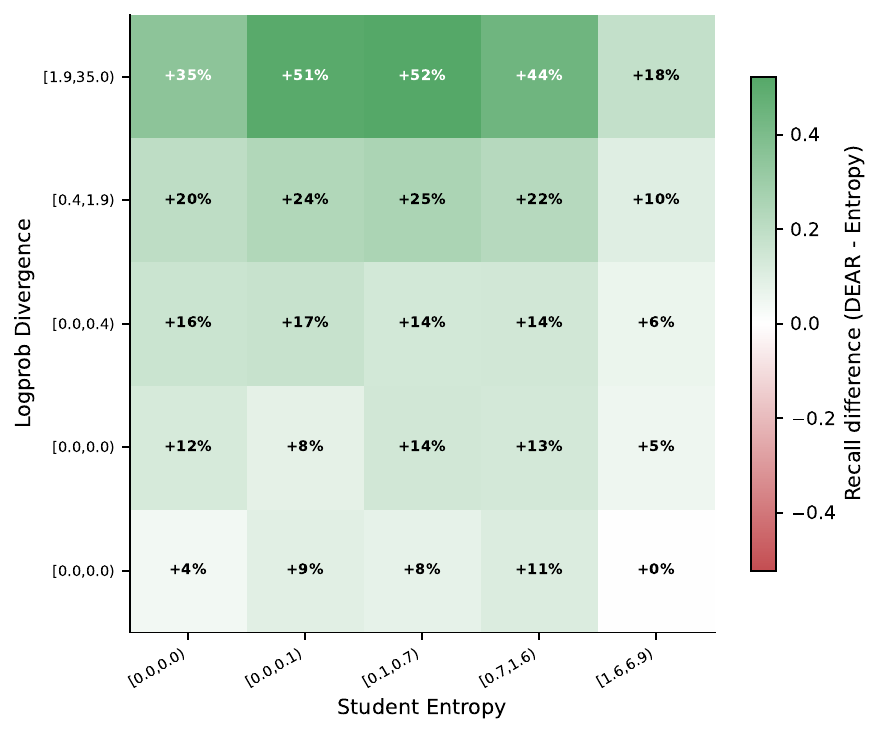}
\caption{\textbf{Evidence recovery targets the correct region.} Each cell shows recall difference (\methodname $-$ decision-only) on the entropy--divergence quantile grid. Gains concentrate in the low-entropy, high-divergence region.}
\label{fig:recall_heatmap}
\end{figure}

The recall-gain heatmap (Figure~\ref{fig:recall_heatmap}) confirms spatially that \methodname's improvement concentrates in the evidence-token region identified in \S\ref{sec:evidence_tokens}.

\paragraph{Evidence discovery is selective.}
Figure~\ref{fig:recovery} shows that \methodname recovers substantially more valuable tokens across all categories, with the largest absolute gain in Math Content.
The Evidence Salience Score ($\attn \times (1 + \divhat)$) discriminates evidence from noise by requiring both semantic connection to decisions and a teacher--student knowledge gap.

\paragraph{Evidence-enriched training signals compound over time.}
\begin{figure}[t!]
\centering
\includegraphics[width=0.85\columnwidth]{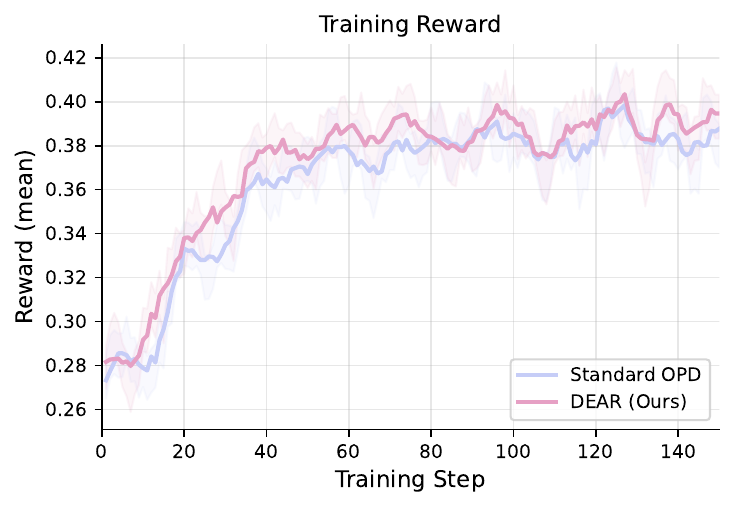}
\caption{\textbf{Training reward over steps.} \methodname achieves higher reward throughout training, with the gap widening over time.}
\label{fig:training_reward}
\end{figure}

Figure~\ref{fig:training_reward} shows that \methodname achieves consistently higher reward, with the gap widening as training progresses.
As the student learns evidence early, its subsequent trajectories improve and expose further evidence tokens, creating a compounding effect.
Standard OPD exhibits elevated clip ratios (Appendix~\ref{app:clip_ratio}), symptomatic of noisy updates from error-accumulated tokens that \methodname avoids.

\subsection{Ablation Studies}
\label{sec:ablation}

We ablate \methodname's components using Setting~C (Qwen3-1.7B distilled from Qwen3-4B-Instruct), which offers the strongest baseline performance and thus the most demanding test for each component's contribution.

\paragraph{Component ablation: decisions alone vs.\ decisions + evidence.}
Removing Stage~2 (evidence discovery) reduces \methodname to entropy-selective OPD (decision-only distillation).
Table~\ref{tab:component_ablation} reports math-average accuracy across all three settings.
Decisions alone perform comparably to standard OPD across settings, confirming the value of selective training even without evidence.
Adding evidence discovery yields a further consistent gain across all three settings, demonstrating that evidence tokens carry complementary knowledge that decisions alone cannot capture.
The gain is largest in Setting~B (cross-family), where the decision--evidence gap is widest.

\begin{table}[t]
\centering
\small
\caption{\textbf{Component ablation.} Math-average accuracy (\%) across all settings. Removing Stage~2 reduces \methodname to decision-only distillation. Evidence discovery provides consistent complementary gains.}
\label{tab:component_ablation}
\begin{tabular}{l ccc}
\toprule
\textbf{Method} & \textbf{Setting A} & \textbf{Setting B} & \textbf{Setting C} \\
\midrule
Standard OPD       & 18.33 & 21.09 & 47.44 \\
Decisions only     & 18.82 & 20.01 & 48.22 \\
\methodname        & \textbf{19.86} & \textbf{21.55} & \textbf{49.06} \\
\bottomrule
\end{tabular}
\end{table}

\paragraph{Sensitivity to decision ratio $p$ and evidence ratio $q$.}

\begin{figure}[t!]
\centering
\includegraphics[width=\columnwidth]{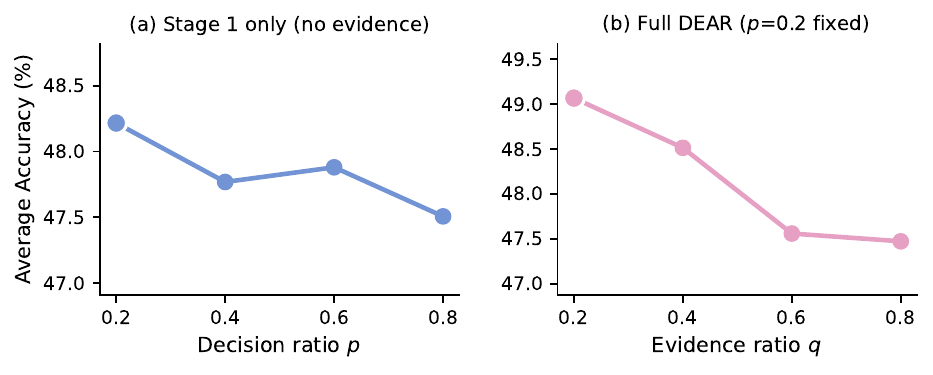}
\caption{\textbf{Sensitivity to selection ratios}. Left: decision ratio $p$ (Stage~1 only, no evidence). Right: evidence ratio $q$ with decisions fixed at $p{=}0.2$. Both stages benefit from sparsity; performance degrades gracefully at higher ratios.}
\label{fig:ablation_ratios}
\end{figure}

Figure~\ref{fig:ablation_ratios} sweeps each ratio independently.
For the decision ratio $p$ (left, Stage~1 only), performance is stable from 0.2 to 0.6 and degrades at $p{=}0.8$, where low-entropy noise tokens dilute the decision set.
This confirms that decision-only distillation is robust to the choice of $p$ within a broad range; the widely-adopted $p{=}0.2$~\citep{wang2026beyond} is near-optimal.
For the evidence ratio $q$ (right, full \methodname with $p{=}0.2$), the optimum is at $q{=}0.2$ with gradual degradation at higher ratios as the cosine gate admits less relevant tokens.
Both stages benefit from sparsity: a focused set of high-quality tokens outperforms broader but noisier coverage.

\section{Related Work}
\label{sec:related_work}

\paragraph{On-policy distillation.}
Knowledge distillation~\citep{DBLP:journals/corr/HintonVD15} transfers capabilities from teacher to student.
OPD extends this to autoregressive generation: the student rolls out under its own policy while the teacher provides token-level supervision~\citep{agarwal2024policy,gu2024minillm}.
Recent analyses identify key failure modes: \citet{luo2026demystifying} trace collapse to repetitive-token advantage amplification; \citet{li2026rethinking} attribute instability to signal degradation on drifted prefixes; \citet{zhang2026fast} show that focusing on the reasoning prefix improves stability.
These studies converge on a common message: indiscriminate training over all tokens is inefficient and destabilizing.
\methodname goes further by asking \emph{which} tokens to select, distinguishing decisions from evidence.

\paragraph{Token-level selective training.}
\citet{wang2026beyond} show that training on the top-20\% highest-entropy tokens matches full-sequence training.
\citet{jin2026entropy} propose forward KL at high-entropy positions; \citet{ko2026scaling} adopt reward-based clipping followed by entropy selection.
These methods establish entropy as an effective proxy for decision tokens.
Concurrently, \citet{xu2026tip} formalize the entropy blind spot, and \citet{feng2024keypoint} propose keypoint-based progressive distillation.
All treat token selection as a single-signal scoring problem.
\methodname instead uses a two-stage process: entropy identifies decisions, then cosine similarity discovers evidence tokens that no single signal can reach.

\paragraph{The structure of reasoning chains.}
Recent work reveals that reasoning ability concentrates at sparse, critical positions.
\citet{shen2026reasoning} show that correcting merely ${\sim}8\%$ of ``decision tokens'' at inference time recovers full reasoning performance, with these tokens strongly enriched for planning-related words.
\citet{wang2026beyond} confirm that high-entropy tokens, which we identify as decision markers, carry the majority of learning signal.
Our work complements this line by identifying a second component: \emph{evidence tokens} that support and justify decisions but are invisible to entropy-based methods due to their low student uncertainty.

\section{Conclusion}
\label{sec:conclusion}

Reasoning chains carry two types of transferable knowledge that cannot be captured by the same selector: decisions, which surface through student uncertainty, and evidence, which hides behind confident errors and must be discovered via its relationship to decisions.
\methodname operationalizes this finding and consistently outperforms standard OPD across math and code benchmarks, with gains scaling with problem difficulty.
Complete reasoning distillation requires transferring not just where to decide, but the knowledge that justifies those decisions.

\clearpage
\section*{Limitations}

\methodname requires access to student hidden states during training, adding memory for one layer of activations per sample.
Our evaluation focuses on mathematical reasoning and code generation; the generality of the decision--evidence structure to other reasoning domains such as commonsense reasoning and scientific QA remains to be verified.
The decision and evidence ratios ($p$, $q$) are fixed throughout training; adaptive scheduling that increases evidence coverage as training progresses may yield further gains.
Our semantic classification of tokens into decisions vs.\ evidence relies on surface-level heuristics. A more principled characterization via causal intervention on individual positions would strengthen the theoretical grounding of the framework.

\bibliography{custom}

\clearpage
\appendix

\section{Why Cosine Similarity Detects Evidence Membership}
\label{app:cosine_analysis}

In causal transformers, the hidden state at position $t$ after $L$ layers, $\hiddenstate_t^L$, is not merely a representation of token $y_t$ but the model's compressed working memory encoding all causally accessible context~\citep{elhage2021mathematical}.
We argue that cosine similarity between deep hidden states captures evidence membership, i.e., whether a token participates in the same reasoning step as a decision, rather than surface structural similarity.
Three independent lines of argument support this:

\textbf{Output logits equivalence.}
Since $\text{logits}_t = \hiddenstate_t^L \cdot W_{\text{unembed}}$, two positions with $\cossim(\hiddenstate_i^L, \hiddenstate_j^L) \approx 1$ produce nearly identical next-token distributions, meaning they occupy the same ``predictive state.''
An evidence token supporting a decision shares the decision's predictive context: both are ``thinking about'' the same intermediate result.

\textbf{Linear representation hypothesis.}
Concepts are encoded as linear directions in hidden-state space~\citep{park2023linear}.
In a $d$-dimensional space (e.g., $d{=}2048$), structural role occupies a small subspace while reasoning content dominates.
Two ``therefore'' tokens in different reasoning chains have dissimilar hidden states because they encode different intermediate results. In contrast, a ``therefore'' and an intermediate step within the same derivation have high similarity because they share reasoning context.

\textbf{Empirical evidence from probing.}
Probing studies~\citep{tenney2019bert} confirm that deep-layer representations are dominated by semantic and functional information, with syntactic features attenuating.
While these findings originate from encoder models, subsequent work on autoregressive LLMs has confirmed that deep layers similarly encode task-relevant semantics over surface form.
This means cosine similarity at deep layers reflects reasoning-content sharing, not surface structural matching.

\section{Semantic Token Classification}
\label{app:semantic_classification}

This appendix documents how response tokens are classified into the three semantic categories---\textbf{Math Content}, \textbf{Reasoning Flow}, and \textbf{Structural}---reported in Figure~\ref{fig:semantic} and Figure~\ref{fig:recovery}.
The classifier is a deterministic, rule-based lexical tagger; it introduces no learned parameters or external model, so the category assignments are fully reproducible.

\paragraph{Categories.}
The three categories are intended to separate the reasoning \emph{skeleton} from its \emph{substance}:
\begin{itemize}[nosep,leftmargin=*]
    \item \textbf{Reasoning Flow}: tokens that steer the logical structure of the derivation---causal and inferential connectives (\textit{therefore}, \textit{since}, \textit{thus}), contrast markers (\textit{however}, \textit{instead}), step markers (\textit{first}, \textit{next}, \textit{finally}), reasoning verbs (\textit{assume}, \textit{prove}, \textit{conclude}), and self-correction cues (\textit{wait}, \textit{actually}). These align with the decision tokens surfaced by entropy.
    \item \textbf{Math Content}: tokens that carry mathematical substance---numbers and numeric expressions, single-letter and subscripted variables, operators and \LaTeX{} fragments, math operation verbs (\textit{substitute}, \textit{simplify}, \textit{factor}), and math-domain nouns (\textit{equation}, \textit{polynomial}, \textit{remainder}). These are the intermediate-step tokens that constitute evidence.
    \item \textbf{Structural}: formatting, punctuation, and function words---articles, copulas, prepositions, pronouns, and generic non-math verbs and nouns.
\end{itemize}

\paragraph{Word-level grouping.}
Classifying raw BPE tokens directly would mislabel subword fragments (e.g., the continuation piece of a split word).
We therefore first group BPE tokens into words using the word-boundary marker (\texttt{\textbackslash u0120}), classify each \emph{word}, and let all of its constituent BPE tokens inherit the word's label.
Pure-operator and pure-digit continuation pieces with no leading space are treated as standalone words, so that an expression such as \texttt{b\^{}2} is split into \texttt{b}, \texttt{\^{}}, \texttt{2} rather than merged.

\section{Training Dynamics: PPO Clip Ratio}
\label{app:clip_ratio}

\begin{figure}[h]
\centering
\includegraphics[width=0.85\columnwidth]{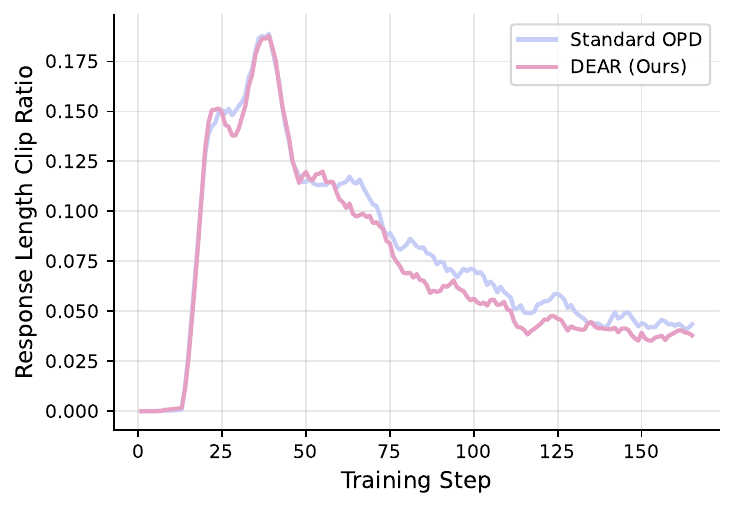}
\caption{\textbf{PPO clip ratio over training.} \methodname maintains a lower clip ratio than standard OPD, indicating more stable policy updates.}
\label{fig:clip_ratio}
\end{figure}

\section{Knowledge-Gap Scoring Ablation}
\label{app:scoring_ablation}

\begin{table}[h]
\centering
\small
\caption{\textbf{Ablation of knowledge-gap scoring functions} on Setting~A (Qwen2.5-1.5B distilled from Qwen2.5-14B). ``Relevance'' = cosine similarity gate $\attn$; ``Gap'' = divergence boost $\divhat$.}
\label{tab:hybrid_ablation}
\resizebox{\columnwidth}{!}{%
\begin{tabular}{@{}lccccccc@{}}
\toprule
\textbf{Scoring} & \textbf{MATH} & \textbf{Minerva} & \textbf{AMC} & \textbf{Olympiad} & \textbf{AIME25} & \textbf{AIME24} & \textbf{Avg.} \\
\midrule
Relevance only ($\attn$) & 52.23 & 12.78 & 18.12 & \textbf{29.37} & 0.83 & \textbf{4.58} & 19.65 \\
Full ($\attn \!\times\! (1{+}\divhat)$) & \textbf{53.15} & \textbf{13.83} & \textbf{18.53} & 29.06 & \textbf{1.25} & \textbf{4.58} & \textbf{20.07} \\
Gap only ($\divhat$) & 52.35 & 13.32 & 18.18 & \textbf{29.37} & \textbf{1.25} & 1.67 & 19.36 \\
\bottomrule
\end{tabular}%
}
\end{table}

The full scoring function ($\attn \times (1 + \divhat)$) outperforms both ablated variants.
Relevance-only selection recovers evidence tokens that are semantically related to decisions but may already be well-aligned; it severely underperforms on AMC and Olympiad where the knowledge gap is large.
Gap-only selection recovers tokens with large teacher--student divergence regardless of reasoning context; it degrades on AIME~2024, confirming that divergence without the cosine gate admits noise from error accumulation.
The multiplicative combination ensures both conditions are met: semantic connection to a decision \emph{and} a knowledge gap worth closing.

\section{Derivation of PG-Style OPD}
\label{app:opd_derivation}

We derive the per-token advantage used in PG-style on-policy distillation (Eq.~\ref{eq:advantage} in the main text).
Let $\piS$ denote the student policy and $\piT$ the teacher.
The OPD objective minimizes the reverse KL divergence between the student and teacher on student-generated trajectories:

\begin{equation}
  \begin{aligned}
    \mathcal{J}_{\mathrm{OPD}}(\theta)
    &= \min_{\theta}\,
    \mathbb{E}_{\prompt \sim D,\, \respond \sim \piS(\cdot \mid \prompt)} \\
    &\quad\left[
      \mathcal{D}_{\mathrm{KL}}\!\left(
        \piS(\respond \mid \prompt)
        \,\|\,
        \piT(\respond \mid \prompt)
      \right)
    \right].
  \end{aligned}
\end{equation}

Expanding the KL divergence and applying the autoregressive chain rule:

\begin{equation}
  \begin{aligned}
    \mathcal{J}_{\mathrm{OPD}}(\theta)
    &= \mathbb{E}_{\prompt,\, \respond \sim \piS}
    \!\left[
      \log \frac{\piS(\respond \mid \prompt)}{\piT(\respond \mid \prompt)}
    \right] \\
    &= \mathbb{E}_{\prompt,\, \respond \sim \piS}
    \!\left[
      \sum_{t=1}^{T}
      \log \frac{\piS(y_t \mid \prompt, y_{<t})}
               {\piT(y_t \mid \prompt, y_{<t})}
    \right].
  \end{aligned}
\end{equation}

Taking the gradient with respect to $\theta$ and applying the log-derivative trick $\nabla_{\theta} \mathbb{E}_{\respond \sim \piS}[f(\respond)] = \mathbb{E}_{\respond \sim \piS}[f(\respond) \nabla_{\theta} \log \piS(\respond)]$:

\begin{equation}
  \begin{aligned}
        &\nabla_{\theta} \mathcal{J}_{\mathrm{OPD}}(\theta)
        = \mathbb{E}_{\prompt \sim D,\, \respond \sim \piS(\cdot \mid \prompt)} \\
        &\left[ \sum_{t=1}^{T}
            \underbrace{\left(
                \log \piS\left(y_{t} \mid \prompt, y_{<t}\right)
                - \log \piT\left(y_{t} \mid \prompt, y_{<t}\right)
            \right)}_{-A_t} \right.\\
        &\quad \left. \cdot \nabla_{\theta} \log \piS\left(y_{t} \mid \prompt, y_{<t}\right)
        \right].
  \end{aligned}
\end{equation}

This takes the standard policy-gradient form $\nabla_\theta \mathcal{J} = \mathbb{E}[\sum_t A_t \nabla_\theta \log \piS(y_t)]$, where the per-token advantage is:

\begin{equation}
    A_t = -\bigl(\log \piS(y_t \mid \prompt, y_{<t}) - \log \piT(y_t \mid \prompt, y_{<t})\bigr).
\end{equation}

This unifies OPD within the reinforcement learning framework: the negative log-probability ratio serves as a dense, token-level reward signal from the teacher.
Intuitively, $A_t > 0$ when the teacher assigns higher probability to token $y_t$ than the student does, encouraging the student to increase that token's likelihood; $A_t < 0$ when the student is already more confident than the teacher, discouraging overcommitment.
In practice, this gradient is approximated via a PPO-style clipped surrogate (Eq.~\ref{eq:opd_loss}), which stabilizes training by bounding the policy update magnitude.

\section{Implementation Details}
\label{app:impl_details}

We implement \methodname based on the VeRL framework~\citep{DBLP:conf/eurosys/ShengZYWZZPL025}.
Tables~\ref{tab:impl_math} and~\ref{tab:impl_code} summarize the training hyperparameters for math and code experiments respectively.

\begin{table}[h]
\centering
\small
\caption{\textbf{Math training hyperparameters} .}
\label{tab:impl_math}
\setlength{\tabcolsep}{4pt}
\begin{tabular}{@{}ll@{}}
\toprule
\textbf{Hyperparameter} & \textbf{Value} \\
\midrule
Training data & DeepMath-103K (level-6) \\
Training epochs & 3 \\
Global batch size & 1024 \\
Micro batch size / GPU & 1 \\
Learning rate & $1 \times 10^{-6}$ \\
LR warmup ratio & 0.0 \\
PPO clip $\epsilon$ & 0.2 \\
Max prompt length & 2048 \\
Max response length & 10240 \\
Max tokens / GPU & 32768 \\
Rollout temperature & 1.0 \\
Rollout top-$p$ & 1.0 \\
Rollout samples ($n$) & 1 (train), 8 (val) \\
KL penalty & None \\
Grad.\ checkpointing & Enabled \\
Inference backend & SGLang (TP=4) \\
GPU mem.\ utilization & 0.8 \\
Hardware & $8\times$ H800 \\
\bottomrule
\end{tabular}
\end{table}

\begin{table}[h]
\centering
\small
\caption{\textbf{Code training hyperparameters} .}
\label{tab:impl_code}
\setlength{\tabcolsep}{4pt}
\begin{tabular}{@{}ll@{}}
\toprule
\textbf{Hyperparameter} & \textbf{Value} \\
\midrule
Training data & Eurus-RL-Code (25K) \\
Training epochs & 5 \\
Global batch size & 1024 \\
Micro batch size / GPU & 1 \\
Learning rate & $1 \times 10^{-6}$ \\
LR warmup ratio & 0.0 \\
PPO clip $\epsilon$ & 0.2 \\
Max prompt length & 2048 \\
Max response length & 10240 \\
Max tokens / GPU & 32768 \\
Rollout temperature & 1.0 \\
Rollout top-$p$ & 1.0 \\
Rollout samples ($n$) & 1 (train), 8 (val) \\
KL penalty & None \\
Grad.\ checkpointing & Enabled \\
Inference backend & SGLang (TP=4) \\
GPU mem.\ utilization & 0.8 \\
Hardware & $8\times$ H800 \\
\bottomrule
\end{tabular}
\end{table}

\section{Training Curves}
\label{app:training_curves}

\begin{figure*}[h]
\centering
\includegraphics[width=\textwidth]{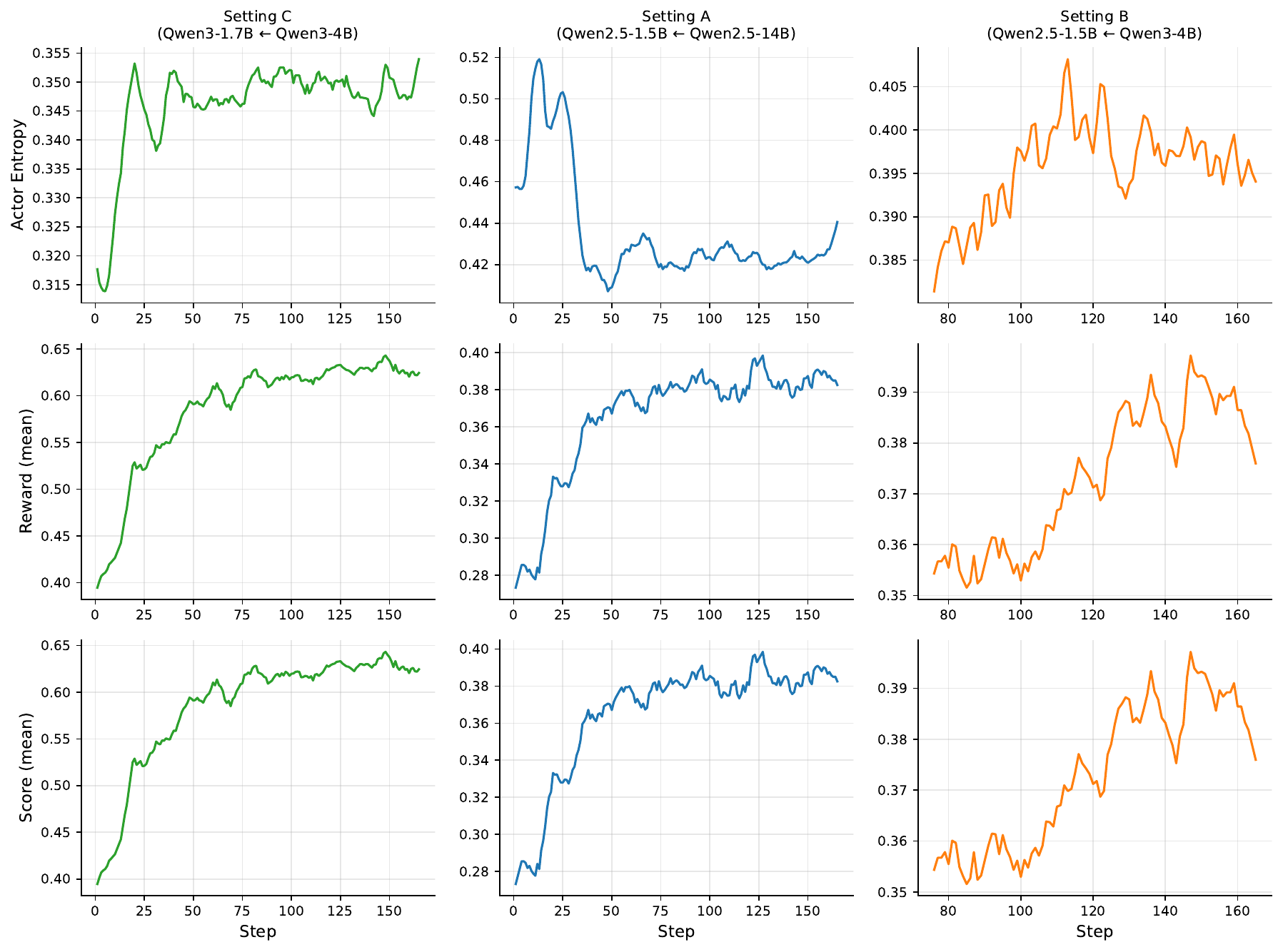}
\caption{\textbf{Training curves on math} (Settings A, B, C). Rows: actor entropy, reward (mean), score (mean). Columns: Setting~C (Qwen3-1.7B $\leftarrow$ Qwen3-4B), Setting~A (Qwen2.5-1.5B $\leftarrow$ Qwen2.5-14B), Setting~B (Qwen2.5-1.5B $\leftarrow$ Qwen3-4B).}
\label{fig:training_curves_math}
\end{figure*}

We report the full training dynamics for all experimental configurations. Figure~\ref{fig:training_curves_math} shows actor entropy, mean reward, and mean score over training steps for the three math settings. Across all configurations, reward and score increase steadily throughout training, confirming stable optimization. Actor entropy decreases as the student becomes more confident, consistent with successful knowledge transfer from the teacher.

\end{document}